\documentclass{article}

\usepackage[tableposition=below]{caption} 

 \usepackage[final]{neurips_2022}

\usepackage[utf8]{inputenc} 
\usepackage[T1]{fontenc}    
\usepackage{hyperref}       
\usepackage{url}            
\usepackage{booktabs}       
\usepackage{amsfonts}       
\usepackage{nicefrac}       
\usepackage{microtype}      
\usepackage{xcolor}         

\usepackage{microtype}
\usepackage{times}
\usepackage{latexsym}
\usepackage{amsmath}
\usepackage{booktabs}
\usepackage{graphicx}
\usepackage{multirow}
\usepackage{bm}

\usepackage{algorithm}
\usepackage{algorithmic}
\usepackage{subfigure}
\usepackage{physics}
\usepackage{bbm}
\usepackage{graphicx}
\usepackage{float}

\usepackage{wrapfig}

\newtheorem{theorem}{Theorem}

\newcommand*{\affaddr}[1]{#1} 
\newcommand*{\affmark}[1][*]{\textsuperscript{#1}}
\title{Gradient Knowledge Distillation for Pre-trained Language Models}

\author{
  Lean Wang\affmark[1], Lei Li\affmark[2], Xu Sun\affmark[1,2] \\
  \affaddr{\affmark[1]School of Electronics Engineering and Computer Science, Peking University }\\
  \affaddr{\affmark[2]MOE Key Lab of Computational Linguistics, School of Computer Science, Peking University}\\
  \texttt{\{lean, xusun\}@pku.edu.cn} \quad \texttt{lilei@stu.pku.edu.cn} \\ 
  } 

\begin{document}
\maketitle
\begin{abstract}
Knowledge distillation~(KD) is an effective framework to transfer knowledge from a large-scale teacher to a compact yet well-performing student. 
Previous KD practices for pre-trained language models mainly transfer knowledge by aligning instance-wise outputs between the teacher and student, while neglecting an important knowledge source, i.e., the gradient of the teacher. The gradient characterizes how the teacher responds to changes in inputs, which we assume is beneficial for the student to better approximate the underlying mapping function of the teacher. Therefore, we propose \emph{Gradient Knowledge Distillation}~(GKD) to incorporate the gradient alignment objective into the distillation process.
Experimental results show that GKD outperforms previous KD methods regarding student performance. Further analysis shows that incorporating gradient knowledge makes the student behave more consistently with the teacher, improving the interpretability greatly.\footnote{Our code is available at \url{https://github.com/lancopku/GKD}.}
\end{abstract}

\section{Introduction}





Knowledge distillation~(KD)~\citep{Hinton2015Distilling,romero2014fitnets} is a classic framework for model compression, which trains a compact student model by utilizing the learned knowledge in a large teacher PLM via teacher-student prediction alignments.
Various alignment strategies have been proposed, like internal representation matching~\citep{Sun2019PatientKD,Sanh2019DistilBERT,Jiao2019TinyBERT} and attention heatmap consistency~\citep{wang2020MiniLM}, and obtain promising efficiency-performance trade-off.
However, previous KD studies for PLMs mostly align the student to the teacher model in a point-to-point manner, neglecting an important knowledge source, i.e., the gradient of the teacher.



Viewing a model as a function mapping the input to the label space, the gradients thus can depict the curve of the function, capturing the model activation changes according to the input perturbation, which we assume can be beneficial for the student model. 
Besides, as gradients are closely related to interpretation, gradient alignment is likely to improve interpretation consistency.

   

Motivated by this, we explore incorporating gradient alignment in knowledge distillation.
However, introducing gradient alignment to KD for PLMS is challenging due to the following two reasons.
First, the inherent discreteness of natural language makes directly computing the deviations w.r.t. the input sentences intractable. 
To remedy this, we instead align the gradient w.r.t the input embedding, and initialize the embedding of the student with that of the teacher and freeze it during training.
Second, our theoretical analysis shows that the commonly adopted Dropout regularization~\citep{Srivastava2014DropoutAS} will disturb the gradient alignment, so we deactivate dropout while conducting distillation. 
Further analysis and experiments can be found in Section~\ref{sec:gkd} and Table~\ref{mainresults}.
Experiments on large-scale datasets in the GLUE benchmark~\citep{wang-etal-2018-glue} demonstrate that gradient knowledge improves the student's performance and behavior consistency with the teacher model.
Besides, our analysis of word saliency~\citep{Ding2019SaliencydrivenWA} shows that the alignments on gradients can benefit the behavior consistency in word saliency level.

\begin{figure}[t!]
 \begin{center}
    \includegraphics[width=0.8\textwidth]{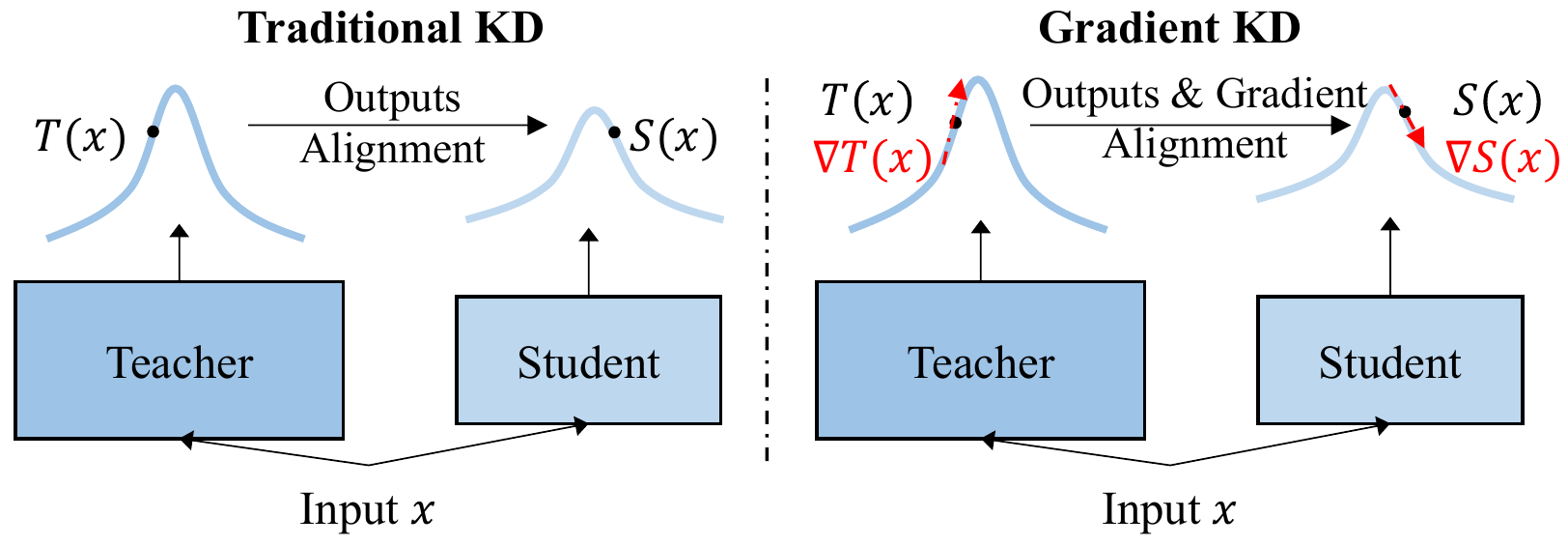}
      \caption{Comparison between traditional KD and our gradient KD. Gradient KD introduces an extra gradient alignment objective to better inform the student of the model behavior of the teacher.
  } 
  \end{center}
\end{figure}

\section{Related Work}
Knowledge distillation has been widely used in computer vision~(CV)~\citep{kdcv,videokd}, natural language process~(NLP)~\citep{Sanh2019DistilBERT,textgenkd} and multimodal field~\citep{multimodalkd,multimodalkd2}.
In the NLP field, knowledge distillation methods can be roughly classified into one-stage methods and two-stage ones. One-stage methods perform distillation at the fine-tuning stage. BiLSTM$_\textrm{SOFT}$~\citep{Tang2019DistillingTK} performs knowledge distillation with the teacher's logits on an augmented dataset, BERT-PKD~\citep{Sun2019PatientKD} performs knowledge distillation with the teacher's logits and hidden states, and PD~\citep{turc2019well} uses a small pre-trained masked language model as a student to enhance the effect of distillation. Two-stage methods perform distillation at both the pre-training stage and the fine-tuning stage. DistilBERT~\citep{Sanh2019DistilBERT} and MobileBERT~\citep{Sun2020MobileBERTAC} focus on the pre-training stage, aiming to get a task-agnostic model that can be fine-tuned or distilled on downstream tasks. TinyBERT~\citep{Jiao2019TinyBERT} first distills a task-agnostic model during pre-training and then performs task-specific distillation on an augmented dataset to further improve performance.

Besides, in the computer vision field, aligning gradients~\citep{Jacobian} or using gradients as a weighting factor~\citep{SCKD} have been explored before. But there is little exploration for PLMs. \citeauthor{NewsBERT} incorporates gradients via a momentum distillation method which is similar to the momentum mechanism~\citep{momentum}, but as far as we know, direct gradient alignment has not been explored in the context of KD for PLMs.


\section{Gradient Knowledge Distillation For PLMs} 
\paragraph{Backgrounds: KD for PLMs}
Formally, for a classification task, we denote $\mathcal{D} = \{\mathbf{x}_i,\mathbf{y}_i\}_{i=1}^N$ as the training dataset with $N$ instances, where $\mathbf{x}_i$ is the input sentence and $\mathbf{y}_i$ is the label. 
Without loss of generality, we take BERT as the representative of PLM model, which transforms the sentence $\mathbf{x}_i$ to a contextualized representation $\mathbf{h}_{i} = \text{BERT}(\mathbf{x}_i)$.
A softmax layer with a learnable parameter $\mathbf{W}$ is appended for producing probability vector $ \mathbf{p}_i=\operatorname{softmax} (\mathbf{W}\mathbf{h}_{i})$ over the label set.
We denote $\mathbf{p}^T_i$ and $\mathbf{p}^S_i$ as the probability vector generated by the teacher $T$ and the student $S$ for the $i$th input, respectively.
KD aims to transform the large teacher's knowledge into a smaller student model. Vanilla KD~\citep{Hinton2015Distilling} achieves it by utilizing the soft targets produced by the teacher:
\begin{align}
\small 
    \mathcal{L}_{\text{KD}} &=(1-\alpha)\mathcal{L}_{\text{CE}}+\alpha \mathcal{L}_{\text{Soft-CE}} \\ 
    \mathcal{L}_{\text{Soft-CE}} &={\tau^2}\sum\limits_{i=1}^N D_{\mathrm{KL}}\left(\mathbf{p}^T_{i,\tau}\| \mathbf{p}^S_{i,\tau}\right),
\end{align}
where $\mathcal{L}_{\text{CE}}$ is the cross-entropy objective for classification tasks, $\mathbf{p}_{i,\tau} =\operatorname{softmax} (\mathbf{W}\mathbf{h}_{i}/\tau)$, $\alpha$ is a hyper-parameter and $\tau$ is the temperature hyper-parameter.
Further explorations introduce alignments on hidden states~\citep{Sun2019PatientKD,Sanh2019DistilBERT} or attention heatmap~\citep{wang2020MiniLM}, and explore dynamic learning schedule~\citep{li-etal-2021-dynamic}.

\paragraph{Improving KD with Gradient Alignments}
\label{sec:gkd}


Different from previous KD studies which consider aligning the teacher and the student in a point-wise manner, we propose to align the change of the model around the inputs by introducing an extra objective on gradient consistency between the student and the teacher. 
In this way, the student can better understand how the output should change when the input changes, which is helpful for the student to behave similarly to the teacher. However, we find that the discreteness of the text and the dropout regularization hinder the alignment of the gradients.
First, as the input sentence tokens are discrete, it is intractable to get the derivative w.r.t the original input sentence.
To remedy this, we instead take the derivative w.r.t the input embeddings. 
Besides, to make sure that the input spaces of the student and the teacher are aligned for calculating the gradient, we initialize the student embedding with that of the teacher, which is a commonly adopted practice in previous KD studies~\citep{Sun2019PatientKD,Jiao2019TinyBERT}, and fix the student embedding layer during the training.
 Denote $p_{m,i}^S$ and $p_{m,i}^T$ as the maximum probability in the obtained class probability distribution from the student and the teacher for an input sentence $\mathbf{x}_i$, the gradient alignment is defined as a mean-square error objective between two normalized deviations:
\begin{equation}
    \mathcal{L}_{\text{GKD}} = \sum\limits_{i=1}^N \sum\limits_{j=1}^{L_i} \Vert \frac{\frac{\partial p_{m,i}^S}{\partial \mathbf{E}^S_{i,j}}}{\Vert \frac{\partial p_{m,i}^S}{\partial \mathbf{E}^S_{i,j}}\Vert_2}
    -
    \frac{\frac{\partial  p_{m,i}^T}{\partial \mathbf{E}^T_{i,j}}}{\Vert \frac{\partial  p_{m,i}^T}{\partial \mathbf{E}^T_{i,j}}\Vert_2}
    \Vert_2^2 \ ,
\end{equation}
where $L_i$ denotes the length of the $i$th tokenized sentence and $\mathbf{E}_{i,j}$ denotes the input embedding vector corresponding to the $j$th token of the $i$th sentence.
As the gradient computation will take all the weights of the model into consideration, the objective encourages the weight updates towards making the student produce consistent gradients with the teacher model.

Second, we find the commonly adopted Dropout~\citep{Srivastava2014DropoutAS} will cause a biased gradient estimation. We provide a mathematical analysis in Theorem~\ref{thm:drop}. Without loss of generality, in Theorem~\ref{thm:drop}, we derive the gradient when inputs are applied with a dropout mask once.

\begin{theorem}
\label{thm:drop}
Consider a function $f$ whose input $\mathbf{x_0}\odot \boldsymbol{\xi}$ comes from a dropout layer, where each component of  $\boldsymbol{\xi}\in\{0,\frac{1}{1-\delta}\}^d$ is drawn independently from a scaled Bernoulli($1-\delta$) random variable. Suppose $f$ can be estimated by its second-order Taylor expansion around $\mathbf x_0$, then
\begin{equation}
    \mathbf{E}_{\boldsymbol \xi}[\nabla f(\mathbf x_0\odot\boldsymbol \xi)] = \nabla f(\mathbf x_0) + \frac{\delta}{1-\delta} \operatorname{diag}( \nabla^2 f(\mathbf x_0))\mathbf x_0 \ .
\end{equation}
\end{theorem}



As the dropout in the teacher model is deactivated, we will align $\nabla T(\mathbf x_0)$ with $\nabla S(\mathbf x_0) + \frac{\delta}{1-\delta} \operatorname{diag}( \nabla^2 S(\mathbf x_0))\mathbf x_0$ when conducting distillation with dropout activated in the student model.
Besides, this bias accumulates when the model goes deeper and leads to more deviated gradients for shallow layers, which is verified in Appendix~\ref{apx:drop}. To remedy this, we deactivate the dropout in the student model for a consistent gradient calculation. 
%
%
%

With the deactivated dropout and the gradient alignment objective, our gradient knowledge distillation~(GKD) is achieved by minimizing the combined objectives ($\alpha$ and $\beta$ are hyper-parameters):
\begin{equation}
    \mathcal L = (1-\alpha) \mathcal L_\textrm{CE}+\alpha \mathcal L_\textrm{soft-CE} + \beta \mathcal L_\textrm{GKD}\ ,
\end{equation}

Furthermore, we apply the gradient alignment on the \texttt{[CLS]} vector to enhance the effect, as \texttt{[CLS]} vectors are usually adopted as the sentence representation for sentence-level prediction:
\begin{equation}
 \mathcal L_\textrm{GKD-CLS} = 
    \sum\limits_{i=1}^N \sum\limits_{j=1}^{m} \Vert \frac{\frac{\partial p_{m,i}^S}{\partial \mathbf{h}^s_{i,S_{j},\textrm{CLS}}}}{\Vert \frac{\partial p_{m,i}^S}{\partial \mathbf{h}^s_{i,S_{j},\textrm{CLS}}}\Vert_2}
    -
    \frac{\frac{\partial  p_{m,i}^T}{\partial \mathbf{h}^T_{i,T_{j},\textrm{CLS}}}}{\Vert \frac{\partial  p_{m,i}^T}{\partial \mathbf{h}^T_{i,T_{j},\textrm{CLS}}}\Vert_2}
    \Vert_2^2 \ ,
\end{equation}
where $\mathbf{h}^t_{i,k,\text{CLS}}$ denotes to the hidden state of the \texttt{[CLS]} token for the $i$th example in the $k$th layer. 
The Skip layer mapping strategy is adopted for $S_{j}$ and $T_{j}$.\footnote{For a 6-layer student, $S_{j}$ takes value $\{1,2,3,4,5\}$, $T_{j}$ takes value $\{2,4,6,8,10\}$ and $m=5$.}
As the \texttt{[CLS]} representations in the student and the teacher can be different, which may hinder gradient alignment, we incorporate an extra loss to align them: 
\begin{equation}
     \mathcal L_\textrm{PKD} = \sum\limits_{i=1}^N \sum\limits_{j=1}^{m} \Vert \frac{\mathbf{h}^S_{i,S_{j},\textrm{CLS}}}{\Vert \mathbf{h}^S_{i,S_{j},\textrm{CLS}}\Vert_2}
    -
    \frac{ \mathbf{h}^T_{i,T_{j},\textrm{CLS}}}{\Vert  \mathbf{h}^T_{i,T_{j},\textrm{CLS}}\Vert_2}
    \Vert_2^2  \ .
\end{equation}
Therefore, the objective of the enhanced version GKD-CLS is ($\alpha$, $\beta$ and $\gamma$ are hyperparameters)
\begin{equation}
      \mathcal L = (1-\alpha)\mathcal  \mathcal L_\textrm{CE}+\alpha \mathcal L_\textrm{soft-CE} +\beta \mathcal L_\textrm{PKD} + \gamma  (\mathcal L_\textrm{GKD} + \mathcal L_\textrm{GKD-CLS}). 
\end{equation}

\section{Experiments}

In this section, we compare our method with other methods from several aspects. We mainly focus on applying our distillation method in the fine-tuning stage due to the limited computational resources. Still, we do experiments to combine our method with Distilbert~\citep{Sanh2019DistilBERT}, to demonstrate that our method can improve the performance of the models distilled in the pre-training stage (\S~\ref{sec:analysis}).

\subsection{Experimental Settings}
\label{sec:experimental_settings}

\paragraph{Datasets}
As fine-tuning the BERT model on small datasets like RTE~\citep{bentivogli2009rte} and MRPC~\citep{dolan2005mrpc} can be quite unstable~\citep{RevisitingFew-sampleBERTFine-tuning,OntheStabilityofFine-tuningBERT}, 
we select four large-scale sentiment classification and natural language inference datasets from the GLUE benchmark~\citep{wang-etal-2018-glue} for more stable evaluation, including \linebreak Stanford Sentiment Treebank~(SST-2)~\citep{socher2013sst}, Quora Question Pairs~(QQP), \linebreak Multi-Genre Natural Language  Inference~(MNLI)~\citep{williams2018mnli} and Question-answering NLI~(QNLI)~\citep{rajpurkar2016squad}. 


\paragraph{Baselines}  
We compare our method with direct fine-tuning and two task-specific KD methods.
Vanilla KD~\citep{Hinton2015Distilling} aligns the logits of the student and the teacher. 
BERT-PKD~\citep{Sun2019PatientKD} utilizes the hidden states of the \texttt{[CLS]} token as well as the logits. Besides, in the first part of Section~\ref{sec:analysis}, we combine our method with DistilBERT~\citep{Sanh2019DistilBERT}, which performs distillation during pre-training.

\paragraph{Training Details}
We use a 12-layer BERT-base-uncased model as the teacher, and the students have the same architecture as the teacher but with $6$ layers. We fine-tune the teacher on specific tasks to get the task-specific teacher. 
We conduct a hyper-parameter search for the baseline methods and our method in a way similar to~\citet{Sun2019PatientKD}. Appendix~\ref{apx:hyperparameter} gives detailed experimental settings.



\subsection{Main Results}

\begin{table}[t!]
\centering
\caption{
Results from the GLUE evaluation server with the best scores in bold. 
The metric for QQP is F1 score and others are accuracy. 
Avg. denotes the average score over SST-2, QQP, MNLI-m, MNLI-mm and QNLI.
}

\begin{tabular}{@{}l|cccccc@{}}
\toprule 
\textbf{Model} & \textbf{SST-2} &  \textbf{QQP} & \textbf{MNLI-m~/~mm} & \textbf{QNLI} & \textbf{Avg.} \\
\midrule 
BERT$_{\text{BASE}}$ & 93.7  & 71.5 & 84.8 / 83.8 & 91.3 & 85.0\\
\midrule 
Fine-tuning & 91.0 &  69.6 & 81.3 / 79.5 & 87.3 & 81.7\\
Vanilla KD & 92.0 & 71.0 & 82.2 / 81.2 & 88.9 & 83.1 \\
BERT-PKD & 91.8 & 71.0 & 82.4 / 81.8 & 89.1 & 83.2\\
\midrule 
GKD & 92.0 & 71.5 & \textbf{82.9} / 81.7 & 89.3 & 83.5 \\
GKD-CLS &  \textbf{93.0} &  \textbf{71.6} & 82.6 / \textbf{81.9} & \textbf{89.5} & \textbf{83.7} \\
 \ w/ Dropout &  91.8 & 71.0 & 82.3 / 81.6  & 89.1 & 83.2\\
\bottomrule
\end{tabular}
\label{mainresults}
\end{table}

\begin{table}[t!]
\centering
 \caption{
Results obtained by fine-tuning or distilling DistilBERT model on specific tasks. 
$^\dag$ denotes the results taken from~\citet{Jiao2019TinyBERT}. 
  Detailed experimental settings can be found in Appendix~\ref{apx:sub_hyperparameter}. 
}
\begin{tabular}{@{}l|cccccc@{}}
\toprule 
\textbf{Model} & \textbf{SST-2} &  \textbf{QQP} & \textbf{MNLI-m/mm} & \textbf{QNLI} & \textbf{Avg.} \\
\midrule 
BERT$_{\text{BASE}}$ & 93.7  & 71.5 & 84.8 / 83.8 & 91.3 & 85.0\\
\midrule 
DistilBERT$^\dag$ & 92.5 & 70.1 & 82.6 / 81.3 & 88.9 & 83.1 \\
\ \ + Vanilla KD & 92.8 &  70.4 & 83.1 / 81.8 & 90.0 & 83.6\\
\ \ + BERT-PKD & 92.9 & \textbf{71.3} & 83.6 / \textbf{82.8} & 90.1 & 84.1\\
\ \ + GKD-CLS &  \textbf{93.7} &  \textbf{71.3} & \textbf{83.7} / \textbf{82.8} & \textbf{90.2} & \textbf{84.3}\\
\bottomrule
\end{tabular}
\label{distilbert_ret}
\end{table}

Table~\ref{mainresults} presents the test results obtained from the GLUE evaluation server. We find that 
(1) Methods with gradient alignment objectives achieve the best performance, verifying our assumption that gradient information benefits the student model. 
(2) Introducing gradient alignment on \texttt{[CLS]}~(GKD-CLS) further boosts the performance, since \texttt{[CLS]} representations play a significant role in classification. 
(3) De-activating dropout is important. Activating Dropout in GKD-CLS leads to inferior results, which is consistent with our analysis~(Section~\ref{sec:gkd}) that dropout will bias the gradient estimation, thus harming the performance.

\subsection{Analysis}
\label{sec:analysis}
\paragraph{GKD Benefits DistilBERT}
\label{gkddistilbert}
Recently, many knowledge distillation methods~\citep{Sanh2019DistilBERT,Sun2020MobileBERTAC,Jiao2019TinyBERT} have explored knowledge distillation during the pre-training stage to build a task-agnostic model or improve task-specific performance. In these works, knowledge distillation can be performed in two stages---the pre-training stage and fine-tuning stage. Since we lack enough computational resources to perform knowledge distillation at the pre-training stage, we instead utilize the $6$-layer DisilBERT model~\citep{Sanh2019DistilBERT}, which is already distilled at the pre-training stage, to perform further GKD distillation at the fine-tuning stage. Here, as the input embeddings of DistilBERT and BERT-base-uncased do not match, we only align the gradients w.r.t. the [CLS] tokens in the gradient alignment objective of GKD-CLS and do not fix the student input embeddings as the teacher input embedding.
The experimental results are shown in Table~\ref{distilbert_ret}, and the detailed experimental settings can be found in Section~\ref{apx:sub_hyperparameter} in Appendix. The results demonstrate that our GKD method can be integrated into the two-stage distillation procedure and can bring more benefits than other methods.  


\paragraph{GKD Improves Output and Interpretation Consistency}
Traditional metrics like accuracy cannot reflect how alike the teacher
and the student behave, which is of great significance in deployments~\citep{xu-etal-2021-beyond}. 
To measure the output consistency, we adopt \textbf{Label Loyalty~(LL)} and \textbf{Probability Loyalty~(PL)} following \citet{xu-etal-2021-beyond}. 
LL measures the faithfulness between the student predictions and the teacher predictions. 
PL measures the distance between the student predictions and the teacher predictions at the distribution level.

Besides, we are interested in whether GKD benefits the consistency of model interpretability, since model interpretation plays an important role in fields like medicine and finance, and a student loyal to the teacher in interpretation is useful. For this purpose, we propose \textbf{Saliency Loyalty~(SL)}.
Specifically, we calculate the teacher's and the student's word saliency distribution using the Grad method~\citep{Ding2019SaliencydrivenWA}, and the SL is defined as the Pearson correlation coefficient between them.

\begin{figure}
\begin{center}

\centerline{\includegraphics[width=1.02\linewidth]{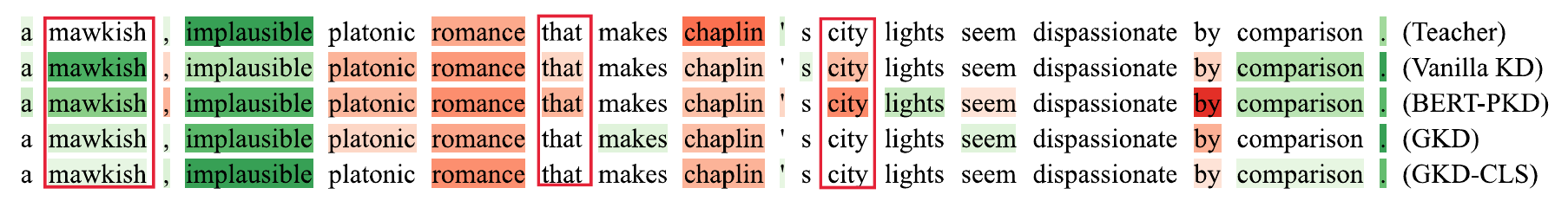}}
\caption{Saliency score visualization on SST-2 test set. Words with positive scores are marked green while negative ones red. While all methods predict the sentence as negative, the word saliency distributions vary. Our method achieves the best consistency with the teacher.}
\label{new_case_study_2}
\end{center}

\end{figure}

We evaluate different methods using these metrics on SST-2 test set. Results are shown in Table~\ref{tab:sst2_loyalty}.\footnote{Results on MNLI can be found in Table~\ref{tab:mnli_loyalty} in Appendix.}
We observe that incorporating gradient into KD improves the behavior consistency regarding all the metrics, especially in SL, i.e. interpretation consistency. 
We also provide a case study of the saliency scores in Figure~\ref{new_case_study_2}, which shows that the saliency scores can vary when all the methods predict the same, and that our methods show the best interpretation consistency, further verifying our motivation.

\begin{table}[t!]
\centering
\caption{Behavior consistency on SST-2 test set. Our GKD methods achieve the best performance on all loyalty metrics, especially on saliency loyalty.
} 

\begin{tabular}{@{}l|ccc@{}}
    \toprule 
     \textbf{Model} &  \textbf{PL} & \textbf{LL} & \textbf{SL}\\
      \midrule 
    BERT$_{\text{BASE}}$ & 100.0 & 100.0 & 100.0\\ 
\midrule 
    Vanilla KD & 93.7 & 93.7 & 31.2 \\
    BERT-PKD & 93.6 & 93.7 & 31.6 \\
    \midrule 
    GKD & 93.9 & 94.3 & 49.7 \\
    GKD-CLS & \textbf{94.9} & \textbf{94.9} & \textbf{53.5} \\
     \bottomrule
    \end{tabular}
\label{tab:sst2_loyalty}
\end{table}

\section{Conclusion}
In this paper, we propose gradient knowledge distillation~(GKD) by matching the unbiased gradient of the student to that of the teacher. 
As the gradient contains higher-order information, the alignment helps the student act more similarly to the teacher.
Experiments show that GKD outperforms baseline methods regarding distillation performance and behavior consistency. 
For future work, we intend to integrate our method with different distillation frameworks to further examine its effectiveness. 



\bibliography{custom}
\bibliographystyle{acl_natbib}

\appendix

\section{Proof of Theorem~\ref{thm:drop}}

\label{proof}
As $f$ can be estimated by its second-order Taylor expansion around $\mathbf x_0$, we can rewrite $f$ as
\begin{equation}
   f(\mathbf x)=f( \mathbf x_{0})+\nabla f( \mathbf x_{0})^T (\mathbf x- \mathbf x_{0})+
    \frac{1}{2}(\mathbf x- \mathbf x_{0})^{T} \nabla^{2} f(\mathbf x_{0})(\mathbf x-\mathbf x_{0})\ ,
\end{equation}
where $\nabla f\left( \mathbf x_{0}\right)$ is a vector and $\nabla^{2} f(\mathbf x_{0})$ is a matrix.

Then, the expectation of gradient is
\begin{align}
\tiny 
    &\mathbf{E}_{\boldsymbol \xi}[\nabla f(\mathbf x_0\odot\boldsymbol \xi)] \nonumber\\
    =&\nabla f(\mathbf x_0) - \nabla^2 f(\mathbf x_0)\mathbf x_0  +  \mathbf{E}_{\boldsymbol \xi}[\boldsymbol \xi \odot \nabla^2 f(\mathbf x_0) (\mathbf x_0\odot\boldsymbol \xi)] \nonumber\\
    =&\nabla f(\mathbf x_0) - \nabla^2 f(\mathbf x_0) \mathbf  x_0 + \nabla^2 f(\mathbf x_0)\mathbf x_0 + \frac{\delta}{1-\delta} \operatorname{diag}( \nabla^2 f(\mathbf x_0))\mathbf x_0 \nonumber\\
    =&\nabla f(\mathbf x_0) +  \frac{\delta}{1-\delta} \operatorname{diag}( \nabla^2 f(\mathbf x_0))\mathbf x_0 \ .
\end{align}

Here, to calculate the expectation, we use the fact that each component of  $\boldsymbol{\xi}\in\{0,\frac{1}{1-\delta}\}^d$ is drawn independently from a scaled Bernoulli($1-\delta$) random variable.

\section{Dropout Biases Gradient}
\label{apx:drop}
To further confirm Theorem~\ref{thm:drop} by experiment, we sample 1000 items from MNLI train dataset to calculate the cosine similarity between the gradient without dropout and the expectation of the gradient with dropout. The model we use is trained by GKD-CLS method. Here, for each item from MNLI, we calculate the gradient $\mathbf v$ 100 times (with dropout activated) and use the average $\frac{\sum_{i=1}^{100} \mathbf v_i}{100}$ to estimate the expectation $\mathbf{E}_{\xi} \mathbf{v}$. The result is shown in Table~\ref{tab:per}.

\begin{table}[H]
    \centering
        \caption{Cosine similarity between the gradient without dropout and the expectation of the gradient with dropout.}
  \begin{tabular}{c|cccccc}
    \toprule
    $\mathbf{v}$ & $\frac{\partial p_{m,i}^s}{\partial \mathbf{E}}$ & $\frac{\partial p_{m,i}^s}{\partial \mathbf{h}_{1,\textrm{CLS}}}$ & $\frac{\partial p_{m,i}^s}{\partial \mathbf{h}_{2,\textrm{CLS}}}$ & $\frac{\partial p_{m,i}^s}{\partial \mathbf{h}_{3,\textrm{CLS}}}$ & $\frac{\partial p_{m,i}^s}{\partial \mathbf{h}_{4,\textrm{CLS}}}$ & $\frac{\partial p_{m,i}^s}{\partial \mathbf{h}_{5,\textrm{CLS}}}$ \\
    \midrule
    $\frac{\mathbf{E}_{\xi}[\mathbf{v}] \cdot \mathbf{v}}{\|\mathbf{E}_{\xi}[\mathbf{v}]\|\cdot \|\mathbf{v}\|}$ &  0.84 & 0.89 & 0.90 & 0.91 & 0.91 & 0.91\\
    \bottomrule
\end{tabular}

    \label{tab:per}
\end{table}

\section{Experimental Settings}
\label{apx:hyperparameter}
\subsection{Settings for the Main Results Shown in 
\label{apx:main_hyperparameter}
Table~\ref{mainresults} }
For hyper-parameter search, we use a strategy similar to the one proposed by \citet{Sun2019PatientKD}. That is, we first search hyper-parameters $\alpha$, $\tau$ for Vanilla KD method, and then fix $\alpha$, $\tau$ to search the additional hyper-parameters in other methods. Besides, we initialize Vanilla~KD~\citep{Hinton2015Distilling}, BERT-PKD~\citep{Sun2019PatientKD} and our method with the embedding layer and first 6 hidden layers of the teacher model.

For all the tasks, we fix the training batch size as 32, the training epoch number as 4, and the learning rate as $5\text{e-}5$.

For Vanilla KD, we perform grid search for $\alpha$ in $\{0.2, 0.5, 0.7\}$ and $\tau$ in $\{5,10,20\}$ and get the best $\alpha, \tau$ on validation set. This $\alpha$, $\tau$ will be used for BERT-PKD, GKD and GKD-CLS. Then, for BERT-PKD, we search $\beta$ in $\{10, 100, 500, 1000\}$; for our GKD method, we search $\beta$ in $\{ 0.05, 0.1,  0.2, 0.4\}$; for our GKD-CLS method, we fix $\beta$ to be $500$ and search $\gamma$ in $\{0.02,0.05, 0.1, 0.2\}$. 

We run all experiments on RTX 2080 Ti GPUs. One signal run of our method takes about 1.5h~/~7.5h~/~7.5h~/~2h for SST-2~/~QQP~/~MNLI~/~QNLI on a single RTX 2080 Ti GPU and the baseline methods take roughly half the time for a single run.

We choose the models that show the highest accuracy on validation datasets to generate the predictions on test datasets, which are then submitted to the official GLUE evaluation server.

\subsection{Settings for the Analysis Results Shown in Table~\ref{distilbert_ret} }
\label{apx:sub_hyperparameter}
The experimental settings for Table~\ref{distilbert_ret} are similar to Section~\ref{apx:main_hyperparameter}, except that we use the DistilBERT model as the student rather than initializing the student from the teacher. Besides, we only align the gradients w.r.t. the \texttt{[CLS]} tokens in the gradient alignment objective of GKD-CLS since the embeddings of DistilBERT and BERT-base-uncased do not match, and we set $\beta$ as $1000$ in GKD-CLS to encourage the alignment of \texttt{[CLS]} tokens. 

\section{Loyalty Results on MNLI}

We also evaluate the label loyalty, probability loyalty and saliency loyalty on the MNLI test dataset. The results are shown in Table~\ref{tab:mnli_loyalty}.
\begin{table}[t!]
    \centering
        \caption{Results of loyalty on MNLI test dataset.} 
    \begin{tabular}{@{}l|ccc@{}}
    \toprule 
     \textbf{Model} &  \textbf{PL}~(m / mm) & \textbf{LL}~(m / mm) & \textbf{SL}~(m / mm)\\
      \midrule 
  BERT$_{\text{BASE}}$ & 100.0~/~100.0 & 100.0~/~100.0 & 100.0~/~100.0\\ 
  \midrule 
    Vanilla KD & 88.2~/~87.4 & 87.9~/~87.0 &  15.8~/~15.9 \\
    BERT-PKD & 88.6~/~87.8 & 88.7~/~87.7 & 17.1~/~16.8 \\
    \midrule 
    GKD & \textbf{89.7~/~89.1} & \textbf{89.5~/~88.9} &  \textbf{35.6~/~35.5} \\
    GKD-CLS & 89.5~/~88.9 & 89.4~/~88.7 & 28.2~/~28.0 \\
     \bottomrule
    \end{tabular}
        \label{tab:mnli_loyalty}
\end{table}

\section{Discussion of Distillation Cost}
\label{sec:distillation_cost}
Since our method needs to align the gradient, we need to conduct backpropagation twice in order to calculate the gradient of  the gradient alignment objective w.r.t. the parameters, which is then used for model parameter update. So, our method requires roughly twice the computational cost as normal KD in the training procedure. For example, on the largest MNLI dataset, our method needs about 7.5h to finish training on a single RTX 2080 Ti GPU while vanilla KD takes about 3.7h. However, the training overhead is negligible as we only need to train the model once for deployments and is totally acceptable considering the better distillation performance. And, the inference speed of our method is the same as normal KD, since both methods use the student of the same architecture for inference.

        
\end{document}